\documentclass[final,nonatbib]{article}

\usepackage[utf8]{inputenc}
\usepackage[T1]{fontenc}

\usepackage{siunitx}
\usepackage{enumitem}

\usepackage{packages/main}
\usepackage[]{packages/neurips}
\usepackage{packages/macros}

\setabbreviationstyle[tool]{long-short}
\newabbreviation[category=tool]{a:tool}{FESTS}{Formally Explainable Spatio-Temporal Scenes}
\newabbreviation[category=tool]{a:strem}{STREM}{Spatio-Temporal Regular Expression Matching}

\setabbreviationstyle[abbreviation]{long-short}
\newabbreviation[category=abbreviation]{a:av}{AV}{Autonomous Vehicle}
\newabbreviation[category=abbreviation]{a:cps}{CPS}{Cyber-Physical System}
\newabbreviation[category=abbreviation]{a:dfa}{DFA}{Deterministic Finite Automata}
\newabbreviation[category=abbreviation]{a:llm}{LLM}{Large Language Model}
\newabbreviation[category=abbreviation]{a:nl}{NL}{Natural Language}
\newabbreviation[category=abbreviation]{a:qspre}{q-SpRE}{Quantified-Spatial Regular Expression}
\newabbreviation[category=abbreviation]{a:re}{RE}{Regular Expression}
\newabbreviation[category=abbreviation]{a:rl}{RL}{Reinforcement Learning}
\newabbreviation[category=abbreviation]{a:spre}{SpRE}{Spatial Regular Expression}
\newabbreviation[category=abbreviation]{a:vlm}{VLM}{Visual Language Model}

\newabbreviation[category=abbreviation]{a:peft}{PEFT}{Parameter-Efficient Fine-Tuning}
\newabbreviation[category=abbreviation]{a:lora}{LoRA}{Low Rank Adaptation}
\newabbreviation[category=abbreviation]{a:ppo}{PPO}{Proximal Policy Optimization}
\glsxtrnewsymbol[description={Alternation}]{s:alternation}{\ensuremath{\mid}}
\glsxtrnewsymbol[description={Data Stream}]{s:datastream}{\ensuremath{D}}
\glsxtrnewsymbol[description={Frame}]{s:frame}{\ensuremath{F}}
\glsxtrnewsymbol[description={Kleene-Star}]{s:kleene-star}{\ensuremath{*}}
\glsxtrnewsymbol[description={Explanations}]{s:explanations}{\ensuremath{E}}
\glsxtrnewsymbol[description={Matches}]{s:matches}{\ensuremath{M}}
\glsxtrnewsymbol[description={Sentences}]{s:sentences}{\ensuremath{S}}
\glsxtrnewsymbol[description={Spatial Logic}]{s:s4u}{\ensuremath{S4_{u}}}
\glsxtrnewsymbol[description={Spatial Term Logic}]{s:s4}{\ensuremath{S4}}
\glsxtrnewsymbol[description={Spatial Formula}]{s:spatial-formula}{\ensuremath{\phi}}
\glsxtrnewsymbol[description={Spatial Regular Expression}]{s:spre}{\ensuremath{Q}}
\glsxtrnewsymbol[description={Environment}]{s:environment}{\ensuremath{\sigma}}

\newglossaryentry{k:qwen}{
    name={Qwen},
    description={The qwen model.}
}
\makeglossaries

\begin{document}

\title{Spatio-Temporal Grounding of Large Language Models from Perception Streams}

%
\author[1]{Jacob Anderson}
\author[1]{Bardh Hoxha}
\author[1]{Georgios Fainekos}
\author[1]{Hideki Okamoto}
\author[1]{Danil Prokhorov}

%
\affil[1]{Toyota Motor North America\authorcr Research \& Development, Ann Arbor, MI, USA}

\maketitle

\begin{abstract}
Embodied-AI agents must reason about how objects move and interact in 3-D space over time, yet existing smaller frontier \glspl{a:llm} still mis-handle fine-grained spatial relations, metric distances, and temporal orderings. 
We introduce the general framework \gls{a:tool} that injects verifiable spatio-temporal supervision into an \gls{a:llm} by compiling natural-language queries into \gls{a:spre} — a language combining regular expression syntax with \gls{s:s4u} spatial logic and extended here with universal and existential quantification. 
The pipeline matches each \gls{a:spre} against any structured video log and exports aligned (query, frames, match, explanation) tuples, enabling unlimited training data without manual labels. 
Training a 3-billion-parameter model on 27k such tuples boosts frame-level F1 from 48.5\% to 87.5\%, matching GPT-4.1 on complex spatio-temporal reasoning while remaining two orders of magnitude smaller, and, hence, enabling spatio-temporal intelligence for Video \gls{a:llm}.
\end{abstract}
\section{Introduction}
\label{sec:introduction}


The ability to comprehend and reason about how a dynamic, three-dimensional world evolves over time is fundamental to embodied AI—spanning household robotics, autonomous driving, and assistive manipulation.
To train and evaluate such systems we also need tooling that can query and annotate spatio-temporal events in video perception logs.
\glspl{a:llm} and \glspl{a:vlm}%
\footnote{“Visual’’ refers to any (potentially multi-modal) model that accepts an image or sequence of images as input.} 
already show promise as task-and-motion planners \cite{ravichandran2024spine,kwon2024language} and low-cost annotators \cite{hirose2024lelan}.
Yet a growing body of work demonstrates that frontier models remain brittle: they mis-judge fine-grained spatial relations \cite{ramakrishnan2024does,tang2024grasp,fu2024blink,majumdar2024openeqa}, lose track of temporal dynamics \cite{ChoiEtAl2025eccv}, and struggle when both aspects matter simultaneously \cite{LiEtAl2025iclr}.
For instance, VLMs often confuse relative object ordering, fail to distinguish identical instances, and cannot reason about metric distance—shortcomings that translate directly into failure modes.

In this paper, we present \gls{a:tool}, a framework that injects rich, verifiable spatio-temporal supervision into an \gls{a:llm}, enabling it to answer -- and explain -- complex video queries.
Our key idea is to leverage \glspl{a:spre} \cite{anderson2023pattern}, a language that fuses regular-expression syntax with S4$_u$ spatial logic, to generate large numbers of self-verifiable queries and corresponding ground-truth matches.
These queries can express properties such as ``{\it find all frames in which a car and a bus start at least 10 m apart and come within 1 m of each other within 20 frames}," which go beyond multiple-choice QA, and naturally scale to 2-D or 3-D data.
Crucially, we extend \glspl{a:spre} to support universal and existential quantification over objects to track entities across time and encode behaviors like ``{\it every pedestrian is at least 1m away from the truck.}”

Recently, Li et al.\ \cite{LiEtAl2025iclr} showed that Video \glspl{a:llm} \cite{PZhangEtAl2025iclrVideoLLM,lin2023video} -- models coupling a video encoder with a language decoder -- can improve reasoning skills through purely textual fine-tuning.
Their evidence suggests that temporal-reasoning bottlenecks lie in the \gls{a:llm} component rather than the video encoder, implying that stronger textual supervision can improve reasoning.
Our framework capitalises on this insight: by generating arbitrarily many \glspl{a:spre}-grounded (query, frames, match, explanation) tuples from any perception dataset, we fine-tune the \gls{a:llm} component to reason about both temporal orderings and spatial relations.

In more detail, given textual video object annotation data which must include object classes and bounding box information, and which may include unique object identifiers, pixel depth information, or other attributes of interest, e.g., color.
Our goal is to fine tune an \gls{a:llm} to be able to reason about arbitrary spatio-temporal patterns which can be encoded with \glspl{a:spre}.
We present a framework which automates query generation and data annotation with the goal of producing any desired size training dataset.
It is important to highlight three benefits of our framework.
First, our framework can be utilized on both real data and artificially generated data. 
Second, and most importantly, with a given video data set or perception data, we can generate an arbitrary number of spatio-temporal queries for training and fine tuning.
Third, our framework can also produce natural language explanations on why a pattern was matched on the annotated dataset.
This additional information can be fed as part of the training process, or even be used in a chain-of-thought spatial reasoning framework as in \cite{liu2025spatialcot}.
To our knowledge, no dataset exists that couples complex queries to spatio-temporal reasoning capabilities of models.
Virtually all the prior works on spatio or spatio-temporal fine tuning use multiple choice question and answering for fine-tuning with much simpler spatio-temporal properties, or they do not explicitly reason about spatial relations \cite{choi2024towards}.

Using our benchmark dataset, we show that with just 27k training examples (each paired with explanations), we boosted a 3-billion-parameter model to be competitive against a state-of-the-art GPT-4.1 model on our training and evaluation dataset. 
This establishes that our framework has the potential to enhance Video \glspl{a:llm} \cite{PZhangEtAl2025iclrVideoLLM,lin2023video} with new spatio-temporal reasoning capabilities since we enable some more complex patterns than \cite{liu2025spatialcot}.
Although our fine-tuned model consistently achieves substantial improvements across varied query complexities and frame lengths, there remains strategic room for further enhancement, particularly in existential queries that involve extended object tracking across frames, where GPT-4.1 currently maintains an advantage.

\paragraph{Contributions}
\label{par:contributions}

Our paper makes the following contributions:

\begin{enumerate}

    \item {\bf Dataset:} We release \gls{a:tool} benchmark dataset, the first automatically-annotated video corpus whose labels are derived from verifiable spatio-temporal queries rather than crowd-sourced labels.
    
    \item {\bf End-to-end pipeline:} \gls{a:tool} ships code to (i) synthesize diverse \gls{a:spre} queries, (ii) match them against structured perception logs, and (iii) export aligned \texttt{(query, frames, match, explanation)} tuples for training or evaluation. 
    
    \item {\bf Pattern matching language extension:} We add existential and universal quantifiers to \gls{a:spre}, enabling persistent object tracking

    \item {\bf Empirical improvements:} Using the resulting ``Query→Explain’’ supervision, we fine-tune a 3B-parameter \gls{a:llm} (Qwen-2.5-Coder-Instruct) from 48.5 \% to 87.5 frame-level F1, keeping competitive with GPT-4.1 on complex spatio-temporal reasoning with orders of magnitude fewer parameters.
    
\end{enumerate}

Collectively, these results show that spatio-temporal fine-tuning, powered by logically-grounded synthetic supervision, can endow LLM  with reasoning skills well beyond what multiple-choice QA alone affords.



\section{Related Work}
\label{sec:literature}

{\bf Spatial reasoning with \gls{a:llm} and \gls{a:vlm}.} 
A series of recent papers show that frontier models still lack spatial reasoning capabilities and propose various model enhancements. 
Chen et al.’s SpatialVLM \cite{chen2024spatialvlm}, Cai et al.’s SpatialBot \cite{cai2024spatialbot}, Cheng et al.’s SpatialRGPT \cite{cheng2024spatialrgpt}, Ma et al.’s 3D-aware SpatialLLM \cite{MaEtAl2025cvpr}, and Zhang et al.’s COMFORT \cite{ZhangEtAl2025iclr} all attempt to patch these gaps with geometric priors or object-centered prompts. 
BLINK \cite{fu2024blink} proposes ``visual" commonsense benchmark problems that humans can answer within seconds, e.g., multi-view reasoning, depth estimation, and reflectance estimation.  
Yet the underlying benchmarks remain limited to local or static relations. 
\gls{a:tool} subsumes this scope by compiling natural-language prompts into \gls{a:qspre} that permit metric constraints, set operations, and universal / existential quantification.

{\bf Spatial benchmarks for \gls{a:llm} and \gls{a:vlm}.} 
The works \cite{tang2024grasp} and \cite{ramakrishnan2024does} propose benchmarks that can evaluate whether frontier models poses spatial intelligence which is natural among animals.  
GRASP \cite{tang2024grasp} demonstrates that cutting edge \gls{a:llm} cannot produce plans given a spatial reasoning problem.
SPACE \cite{ramakrishnan2024does} exposes failures of \gls{a:llm} and \gls{a:vlm} to produce a mental map of the environment when traversing it.
It also demonstrates that foundation models cannot perform smaller-scale reasoning about object shapes and layouts.
\gls{a:tool} has orthogonal goals and evaluation criteria to GRASP and SPACE.
However, it would be interesting to evaluate if \gls{a:tool} can also improve spatial intelligence in frontier models.

{\bf Video-LLM benchmarks and temporal reasoning.}
Temporal understanding has progressed from early captioning datasets to full video-LLM challenges. 
Li et al. \cite{LiEtAl2025iclr} prompt VLMs for temporal localization and reveal poor clip-level accuracy. 
Li et al. \cite{LiEtAl2025iclr} further demonstrate that purely textual fine-tuning lifts ordering performance and temporal localization.
V-STaR benchmark \cite{LiEtAl2025iclr} assesses spatio-temporal reasoning ability in answering questions in the context of ``when", ``where", and ``what".
Mementos \cite{wang2024mementos} stresses sequence reasoning over image sets, while PaLM-E \cite{driess2023palm} proposes and evaluates embodied language models with additional sensing modalities. 
The work in \cite{PZhangEtAl2025iclrVideoLLM} shows that by simply expanding context windows improves performance in performance on long video question-answering benchmarks.
NSVS-TL \cite{ChoiEtAl2025eccv} shows that current \gls{a:vlm} fail at long-term reasoning across
frames and propose a temporal logic based framework for temporal reasoning. 
Nearly all approaches (besides NSVS-TL) produce benchmarks based on multiple-choice labels or short captions and question-answering. 
Even though the aforementioned approaches focus on temporal relations across frames, they do not really consider spatial reasoning at the same fidelity as \gls{a:tool}.
\gls{a:qspre} instead produces verifiable \texttt{(query, frames, match, explanation)} tuples that jointly stress spatial and temporal reasoning, and its generator can wrap any perception log—including the clips used by other benchmarks.


\section{The \glsfmtlong{a:strem} Framework}
\label{new:sec:preliminaries}

This section reviews the \gls{a:strem} framework \cite{anderson2023pattern}, highlights its limitations, and presents our contributions to it.
The \gls{a:strem} framework \cite{anderson2023pattern} is designed to match queries over perception data streams.
The queries are expressed as \glspl{a:spre}, which combine \glspl{a:re} with the spatial logic \gls{s:s4u} \cite{kontchakov2007spatial}, enabling patterns to capture both temporal and spatial relationships among objects.
The matching procedure uses a formal-methods approach based on \gls{a:dfa}, which determines whether a perception stream satisfies a given query.

\subsubsection{Limitations}
\label{new:sec:preliminaries:strem:limitations}

In the current variation, there are several limitations to the \gls{a:strem} framework that do not support the ability to perform more complex temporal queries.

In the current version, \gls{a:spre} queries such as, a simple ``{\it Find all frames where the same pedestrian is present for five frames}'', or more complicated, ``{\it Find all frames where the same pedestrian overlaps with any car or bus for five frames}'' are not possible.
Furthermore, reasoning over all kinds of objects at a specific point in time across multiple points of time is not possible and, thus, queries such as, ``{\it Find all frames where all cars are more than 500 units away from any pedestrian for three frames}'' do not have any inherit support.
These limitations enforce a per-frame reasoning query to be formed by the user and thus does not enable a wide range of multi-frame temporal reasoning expressions that would otherwise strengthen the capabilities of the querying language overall.

\subsection{Adding Quantification Support}
\label{new:sec:preliminaries:quantification}

In order to support spatial object tracking and reasoning across time, we introduce quantifiers as part of our query language.
We first adjust the syntax of the \gls{a:spre} grammar to include \gls{a:re}-level quantifiers.
The modified \gls{a:spre} syntax is 

\begin{equation}
    \label{new:eq:preliminaries:quantification:1}
    \gls{s:spre} \coloneqq
    \gls{s:spatial-formula} \mid
    \gls{s:spre}_{1} \cdot \gls{s:spre}_{2} \mid
    \gls{s:spre}_{1} \gls{s:alternation}\, \gls{s:spre}_{2} \mid
    \gls{s:spre}^{\gls{s:kleene-star}} \mid
    \exists{x}. \gls{s:spre} \mid
    \forall{x}. \gls{s:spre}
\end{equation}

where $\exists$ and $\forall$ correspond to the introduced existential and universal quantifiers, and $x$ is a variable that can be used anywhere in the scope of the operators.
The rest of the operators are standard REGEX operators: concatenation ($\cdot$), which is typically omitted, choice ($\gls{s:alternation}$), and repetition ($^{\gls{s:kleene-star}}$).

Intuitively, concatenation is used to require a sequence of occurrences, e.g., $[car]\cdot[car]$ or simply  $[car][car]$, which states that we are looking for two consecutive frames where in each frame some car is present (not necessarily the same car).
In contrast, when we use the quantifier $\exists$ ({\it exists}), e.g., $\exists x \leftarrow [car] . xx$, then we enforce the car to be the same across the two frames.
Here, we use the notation $x \leftarrow [car]$ to indicate that the variable $x$ while refer to a specific object of type $car$.
Similarly, the expression $\forall x \leftarrow [car] . xx$ queries the perception stream for two frames where if a $n$ cars are present in frame $i$, then the same cars are also present in frame $i+1$. 
In other words, we are looking for two frames, where if a car is present in frame $i$, then it is also present is frame $i+1$.

The choice (or union) operator allows as two select between two choices, e.g., $[car]\,\gls{s:alternation}\,[bus]$ (we are looking for a frame where a car or a bus exists).
Finally, the repetition (Kleene star) operator allows us to ask for a repetition of sequence, e.g., if we are looking for the longest alternating sequence where a in a frame we see a car and in the next frame we see a bus, then we would simply write $([car][bus])^{\gls{s:kleene-star}}$.


{\bf Remark:}
Even though on the surface the change in the syntax may seem simple, the addition of quantifiers require complete redevelopment of the semantics and the corresponding query matching algorithms, accordingly.  

\section{\glsfmtlong{a:tool}}
\label{new:sec:fests}

\begin{figure}[t]
    \centering
    \resizebox{1.0\linewidth}{!}{\begin{tikzpicture}[
    node distance={1cm},
    rect/.style={
        draw,
        rectangle,
        minimum width={2.5cm},
        minimum height={1cm},
        fill={gray!10},
        drop shadow={black},
    },
    trap/.style={
        draw,
        trapezium,
        trapezium stretches={true},
        trapezium left angle={60},
        trapezium right angle={120},
        minimum width={3cm},
        minimum height={1cm},
        fill={gray!10},
        drop shadow={black},
    },
    label/.style={
        draw,
        circle,
        minimum size={0.25cm},
        fill={black},
        text={white},
    }
]

    \node (a) [trap] at (0, 0) {\glsfmtshort{a:spre}};
    \node (b) [trap, below={0.25cm of a}] {Scene};
    \node (c) [rect, right={of a}, minimum width={3cm}] {\textbf{\glsfmtshort{a:strem}}};
    \node (d) [trap, right={of c}, yshift=0.75cm] {Explanations};
    \node (e) [trap, right={of c}, yshift=-0.5cm] {Matches};
    \node (h) [rect, right={of d}] {Naturalizer};
    \node (i) [rect] at ($(h |- (0, -2.5)$) {Packager};
    \node (l) [trap, fill={green!25}, right={of i}] {$(\gls{s:spre}^{\prime}, \gls{s:datastream}^{\prime}, \gls{s:matches}, \gls{s:sentences})$};
    \node (m) [trap, below={0.25cm of b}] {\glsfmtshort{a:nl}};

    \node (j) [border={$(d.west)+(-0.2cm,0)$}{$(d.east)+(0.2cm,0)$}{d.north}{e.south}] {};
    \node (k) [anchor={south west}] at (j.north west) {};

    \node (t) [border={$(a.west)+(-0.25cm,0)$}{$(a.east)+(0.25cm,0)$}{a.north}{m.south}] {};
    \node (u) [anchor={south west}] at (t.north west) {\textbf{Inputs}};

    \draw[->] (a) -- (c);
    \draw[->] (c) -- (j);
    \draw[->] (d) -- (h);
    \draw[->] (h) -- (i);
    \draw[->] (i) -- (l);
    \draw[->] (b) -| (c);
    \draw[->] (e) -| (i);
    \draw[->] (m) -- (i);

    \node (q) [right={0.15cm of h}] {(2)};
    \node (r) [above={0.15cm of c}] {(1)};
    \node (s) [below={0.15cm of i}] {(3)};
\end{tikzpicture}}
    \caption{The \glsfmtshort{a:tool} framework begins with (1) which processes the \glsfmtshort{a:spre} and perception stream inputs to produce the two formal method-based results; (2) processes the explanation to improve readability for \glsfmtshort{a:llm}s; and (3) packages this into a distributable data formats.}
    \label{fig:methodology:dataset:1}
\end{figure}
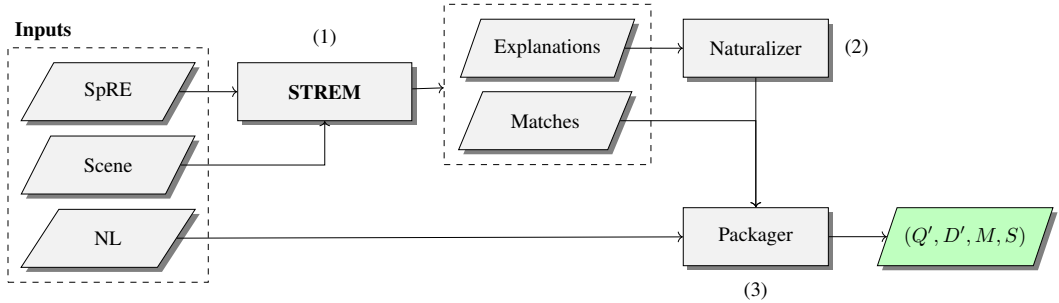

The \gls{a:tool} framework (see \cref{fig:methodology:dataset:1}) accepts as input a data stream \gls{s:datastream} of downstream perception-based data such as object annotations; examples of pre-existing datasets containing such information include Woven Perception \cite{kesten2019woven} or nuScenes \cite{caesar2020nuscenes}.
As output, the \gls{a:tool} data pipeline returns a perception stream of with each entry organized as follows:

\begin{equation}
    (\gls{s:spre}^{\prime}, \gls{s:datastream}^{\prime}, \gls{s:matches}, \gls{s:sentences})
\end{equation}

where $\gls{s:spre}^{\prime}$ is the \gls{a:nl} variant of the \gls{a:spre} query \gls{s:spre}, $\gls{s:datastream}^{\prime} = (\gls{s:frame}_{i}, \gls{s:frame}_{i+1}, \ldots, \gls{s:frame}_{j}) \subseteq \gls{s:datastream}$ is the sampled data stream, \gls{s:matches} is the set of matches from \gls{a:strem}, and \gls{s:sentences} is the set of \gls{a:nl} explanations linearized from the set of explanations \gls{s:explanations}.

Let us consider the following \gls{a:nl} query written for an \gls{a:av} system affixed with image-based sensors and a downstream object detector:

\begin{quote}
  \textit{Find all frames where the bounding box of the same car intersects with a bounding box of a bus for two frames.}
\end{quote}

From this query, the goal is to identify frames from the perception stream that match the properties outlined.
This query is composed of both spatial properties such as \textit{intersection} as well as temporal properties such as \textit{sequences}.
However, while current \glspl{a:llm} such as GPT-4o \cite{hurst2024gpt} initially showcase positive performance on single property-based queries, queries containing a mix of both spatial and temporal elements begin to demonstrate failures.
These failures consist of hallucinations in the perception stream, incorrect ranges, and reduced accuracy over longer traces as concluded in \cite{ChoiEtAl2025eccv}.
To improve upon these limitations, we utilize fine-tuning of \glspl{a:llm} through a formal methods-based approach to the data generated for training and fine-tuning of the models.

\section{Experiments}
\label{new:sec:experiments}

To evaluate the effectiveness of our approach, we fine-tune an \gls{a:llm}, Qwen2.5-3B-Instruct \cite{qwen2025technical}, on the outputs of our framework from an \gls{a:av} perception dataset, Woven Perception \cite{kesten2019woven}.
In the following sections, the dataset composition, fine-tuning procedure, evaluation metrics, and results are presented.

\subsection{Dataset Composition}
\label{new:sec:experiments:dataset}

To fine-tune an \gls{a:llm} on the outputs of our framework, a perception stream source is required.
The Woven Perception \cite{kesten2019woven} dataset was chosen for its comprehensive selection of perception streams and high-quality, hand-labeled object annotations.
This dataset is comprised of 180 different scenes with each scene containing a stream of 126 frames from 7 different monocular camera sensors, which provides 1.2K+ perception streams to process with our framework.

To generate the data for fine-tuning, the perception streams were sampled at incremental frame lengths of 1, 2, 4, 6, 8, 10, 12, 14, and 16 to gradually increase the difficulty for the \gls{a:llm}.
For each sample, our framework joins the satisfaction result and explanation from the \gls{a:strem} framework with the corresponding \gls{a:nl} query and perception stream data for 15 templated queries.
This procedures yields 27K+ outputs as the inputs to fine-tune the \gls{a:llm} on.

\subsubsection{Query Types}
\label{new:sec:experiments:dataset:query}

The queries we fine-tune the model on can be grouped into five distinct categories.
These categories and considerations of each are outlined below:

\begin{enumerate}
    \item \textbf{Sequence}: A query containing multiple temporally adjacent events.
    \item \textbf{Spatial}: A query that contains operations such as intersection of bounding boxes.
    \item \textbf{Temporal}: A query that contains eventual events.
    \item \textbf{Metric}: A query that contains measurement-based operations.
    \item \textbf{Existential}: A query that contains reasoning on the same or all objects over time.
\end{enumerate}

\subsection{Models and Fine-Tuning Configurations}
\label{new:sec:experiments:finetuning}

The fine-tuning was performed entirely on the \gls{a:llm}, Qwen2.5-3B-Instruct \cite{qwen2025technical}.
This model was selected for several reasons: (1) publicly and readily available, (2) small parameter size, (3) ideal for task completion and fine-tuning, and (4) size of context-length.
The model was fine-tuned under the following two training configurations:

\begin{enumerate}[label=C\arabic*.]
    \item \textbf{Supervised Fine-Tuning}: The model was trained exclusively on the \emph{query} and \emph{match} outputs of our framework, with no \emph{explanation} field.
    The \gls{a:peft} using the \gls{a:lora} method was applied to the attention and MLP layers with a rank of 16, scaling of 32, and a dropout of 0.05; trained for 5 epochs with an effective batch size of 60; optimized with AdamW (8-bit) with a learning rate of $1 \times 10^{-5}$ and cosine scheduling.

    \item \textbf{Supervised Fine-Tuning with Reinforcement Learning}: The model was pre-trained from the C1 configuration.
    The \gls{a:rl} with \gls{a:ppo} used where the \gls{a:ppo} used a custom hierarchical-based reward function (see \cref{new:sec:experiments:evaluation}); trained for 1 \gls{a:ppo} epoch with 4 optimization epochs per \gls{a:ppo} batch; optimized with AdamW (8-bit) with a learning rate of $1 \times 10^{-6}$, effective batch size of 4, a KL divergence coefficient of 0.05, and upper bound of 512 tokens for rollouts.
\end{enumerate}

In addition, the fine-tuned models were compared against the GPT-4.1\footnote{This model was accessed and used for evaluation on 05/01/2025.} \cite{achiam2023gpt} model representing the state-of-the-art and the Qwen2.5-Coder-3B-Instruct model \cite{qwen2025technical} representing the baseline.

\subsection{Evaluation Metrics}
\label{new:sec:experiments:evaluation}

To evaluate the model during fine-tuning, we developed two methods distinct for each fine-tuning configuration in \cref{new:sec:experiments:finetuning}.

For the C1 configuration, the causal language modeling objective is optimized using cross-entropy loss, minimizing differences between the predicted and ground-truth token probabilities such that all tokens except the results are masked.

For the C2 configuration, a hierarchical-based reward function is used.
This reward function evaluates several properties including: (1) structural validity such as XML formatting; (2) match accuracy with mAP IoU and exact match; and (3) reasoning fidelity, which assesses semantic similarity to ground-truth explanations using a sentence transformer from \cite{transformers2021all} and numerical IoU of referenced frames.
The penalties of the reward function account for excessive response length, spurious text outside delimited tags, and invalid formats.

While the reasoning fidelity guides the \gls{a:rl} training, it noted that primary performance metrics in \cref{sec:experiments:results} focus on the accuracy of the predicted frames.

\subsection{Results}
\label{sec:experiments:results}

\paragraph{Main Findings.}
\cref{tab:experiments:results:compact} summarizes key results. Adding explanations yields large improvements. Supervised fine-tuning on query–answer pairs (Q-SFT) improves overall Frame~F1 from 48.5\% to 80.4\%. Reinforcement learning on top of those explanations (Q-SFT+RL) pushes Frame~F1 to 87.5\%, just above GPT-4.1 at 84.8\%. The jump is primarily driven by recall (+28~points during SFT) and then by precision (+2.2~points during RL). Exact-match rises from 25.0\% (Baseline) to 56.6\% (Q-SFT) and 64.5\% (Q-SFT+RL).

\begin{figure}[t]
  \centering
  \includegraphics[width=1.0\textwidth]{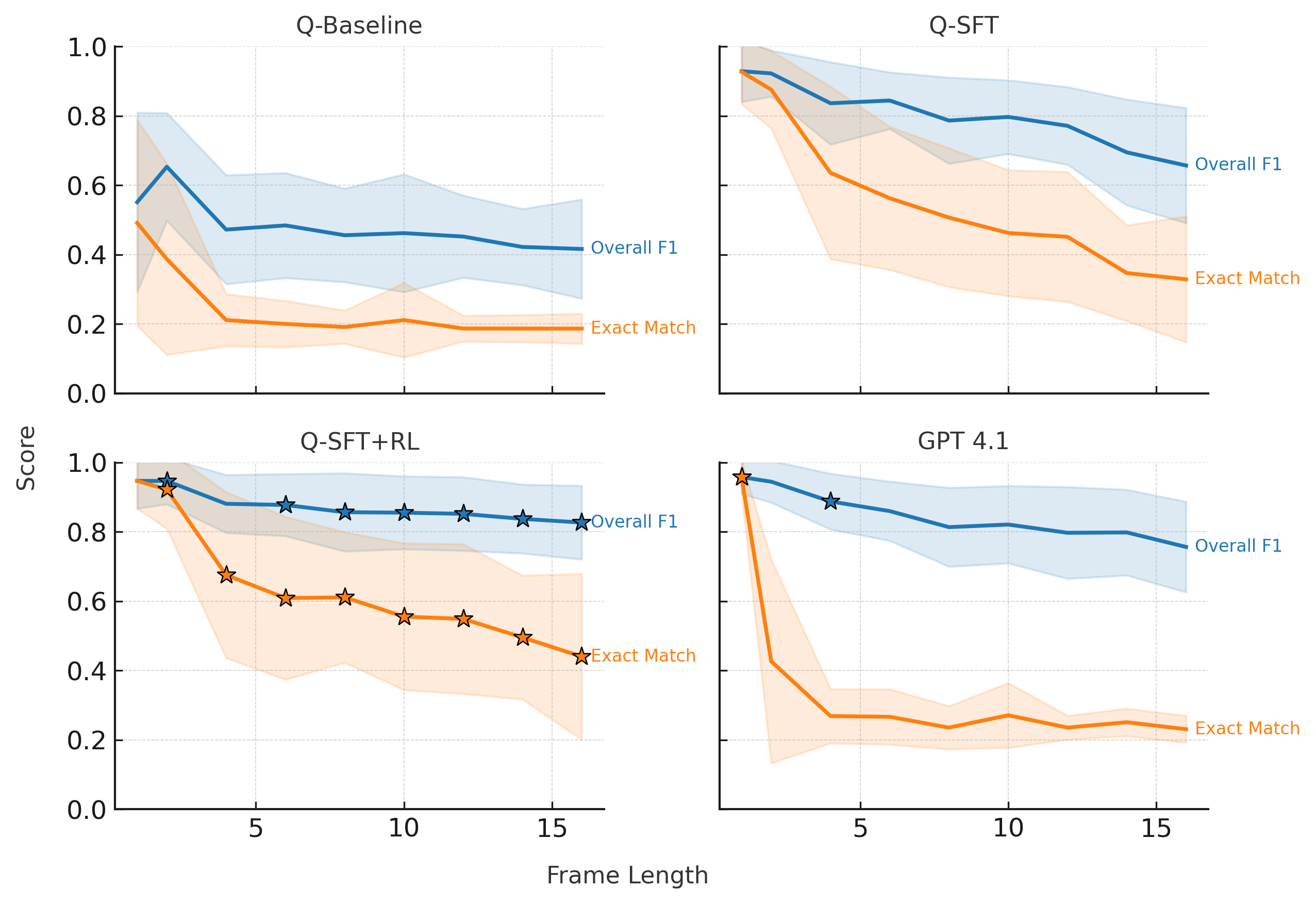}
  \def\svgwidth{1.0\linewidth}
  \vspace{-5pt}
  \caption{Average Performance across frame lengths. Blue = Overall F1, orange = Exact Match; shaded $\sigma$. Starred points denote the best-performing model for each frame length.}
  \label{fig:experiments:results:2}
\end{figure}

\paragraph{Impact of Query Length.}
\cref{fig:experiments:results:2} shows that gains scale with query length. For 16-frame inputs Frame~F1 climbs from 65.7 \% (Q-SFT) to 82.7 \% (Q-SFT+RL), a +17-point jump. Exact-match improves by +39.5~points over baseline and by +29.5~points over GPT-4.1 (64.5 \% vs.\ 35.0 \%). Most of the improvement comes from SFT. RL improves a further 7.9~points.

\begin{table}[htbp]
\centering
\caption{Performance comparison across models and target lengths. Metrics: Frame F1 (F1$_f$), Exact Match (EM), and Segment F1 (F1$_s$). Best results per column are in \textbf{bold}.}
\label{tab:experiments:results:compact}

\scriptsize
\setlength{\tabcolsep}{4pt}
\renewcommand{\arraystretch}{0.95}
\begin{tabular}{@{}lccccccccccccccc@{}}
\toprule
& \multicolumn{3}{c}{4} & \multicolumn{3}{c}{8} & \multicolumn{3}{c}{12} & \multicolumn{3}{c}{16} & \multicolumn{3}{c}{Overall} \\
\cmidrule(lr){2-4} \cmidrule(lr){5-7} \cmidrule(lr){8-10} \cmidrule(lr){11-13} \cmidrule(lr){14-16}
Model & F1$_f$ & EM & F1$_s$ & F1$_f$ & EM & F1$_s$ & F1$_f$ & EM & F1$_s$ & F1$_f$ & EM & F1$_s$ & F1$_f$ & EM & F1$_s$ \\
\midrule
Q-Baseline & 0.472 & 0.211 & 0.325 & 0.456 & 0.191 & 0.258 & 0.452 & 0.187 & 0.231 & 0.416 & 0.187 & 0.215 & 0.485 & 0.250 & 0.326 \\
Q-SFT      & 0.836 & 0.636 & 0.644 & 0.786 & 0.507 & 0.582 & 0.771 & 0.451 & 0.565 & 0.657 & 0.329 & 0.454 & 0.804 & 0.566 & 0.636 \\
Q-SFT+RL   & 0.881 & \textbf{0.676} & \textbf{0.694} & \textbf{0.856} & \textbf{0.611} & \textbf{0.686} & \textbf{0.852} & \textbf{0.549} & \textbf{0.659} & \textbf{0.827} & \textbf{0.440} & \textbf{0.629} & \textbf{0.875} & \textbf{0.645} & \textbf{0.723} \\
GPT-4.1    & \textbf{0.888} & 0.269 & 0.640 & 0.813 & 0.236 & 0.528 & 0.797 & 0.236 & 0.497 & 0.756 & 0.231 & 0.447 & 0.848 & 0.350 & 0.610 \\
\bottomrule
\end{tabular}
\end{table}

\paragraph{Per-Query Type Performance.}
\cref{tab:experiments:results:2} confirms the pattern across five query types. SFT improves F$_1$ by +0.35 (Sequence), +0.35 (Spatial), +0.31 (Temporal), +0.31 (Metric), and +0.28 (Existential). RL improves recall above 0.90 for Sequence, Metric, and Existential and raises F$_1$ to 0.90+ for Sequence, Metric, and Temporal. Spatial and Existential stay at 0.82–0.83 F$_1$. Exact-match is 0.821 for Sequence and 0.706 for Metric but does not perform as well at 0.531 for Existential queries.

\paragraph{Summary.}
The query-explanation–RL pipeline delivers consistent, order-of-magnitude improvements with minimal extra annotation, suggesting strong potential for transfer to other video-language tasks governed by symbolic logic.

Our analysis refrains from comparing against \textit{reasoning} models, as this introduces an additional learning signal and thus another source of variance, complicating attribution in experimental studies. Moreover, reasoning models exacerbate the inherent textual-context limitations of LMMs, necessitating truncated analyses or aggressive token pruning to accommodate input data within memory constraints.

\begin{table}[htbp]
\centering
\caption{Precision ($P$), Recall ($R$), F$_1$, and Exact-Match (EM) for each query type. Values are averaged across 8-, 12-, and 16-frame lengths.}
\label{tab:experiments:results:2}
\resizebox{1.0\textwidth}{!}{
\begin{tabular}{lcccccccccccc}
\toprule
\multirow{2}{*}{Query type} &
\multicolumn{4}{c}{Base} &
\multicolumn{4}{c}{Stage-I: SFT} &
\multicolumn{4}{c}{Stage-II: SFT + RL} \\
\cmidrule(lr){2-5} \cmidrule(lr){6-9} \cmidrule(lr){10-13}
& $P$ & $R$ & F$_1$ & EM
& $P$ & $R$ & F$_1$ & EM
& $P$ & $R$ & F$_1$ & EM \\
\midrule
Sequence     & 0.875 & 0.668 & 0.570 & 0.206 & 0.931 & 0.961 & 0.923 & 0.715 & \textbf{0.971} & \textbf{0.978} & \textbf{0.962} & \textbf{0.821} \\
Spatial      & 0.785 & 0.557 & 0.415 & 0.190 & 0.825 & 0.853 & 0.761 & 0.502 & \textbf{0.845} & \textbf{0.905} & \textbf{0.819} & \textbf{0.552} \\
Temporal     & 0.863 & 0.597 & 0.504 & 0.348 & 0.883 & 0.857 & 0.810 & 0.553 & \textbf{0.910} & \textbf{0.902} & \textbf{0.866} & \textbf{0.615} \\
Metric       & 0.872 & 0.551 & 0.457 & 0.200 & 0.903 & 0.830 & 0.769 & 0.597 & \textbf{0.917} & \textbf{0.960} & \textbf{0.903} & \textbf{0.706} \\
Existential  & \textbf{0.875} & 0.574 & 0.480 & 0.306 & 0.817 & 0.861 & 0.757 & 0.463 & 0.826 & \textbf{0.937} & \textbf{0.828} & \textbf{0.531} \\
\midrule 
Average      & 0.854 & 0.589 & 0.485 & 0.250 & 0.872 & 0.872 & 0.804 & 0.566 & \textbf{0.894} & \textbf{0.936} & \textbf{0.876} & \textbf{0.645} \\
\bottomrule
\end{tabular}}
\end{table}

\section{Limitations}
\label{new:sec:limitations}

The current set of limitations of this work includes: (1) missing component to automate the translation between \gls{a:nl}-based queries into \gls{a:spre} counterparts; (2) the perception stream strictly requires pre-labeled data; (3) accuracy of the results depends on the quality of the sourced labels; (4) manual curation and creation of the queries are required; and (5) evaluation on a single model.

\section{Conclusions}
\label{new:sec:conclusions}


In this work, we developed \gls{a:tool}, a benchmark dataset leveraging verifiable, logically-grounded queries for automated annotation and explainability, substantially advancing video-language model training without reliance on crowd-sourced labels. 
Through systematic fine-tuning of the Qwen-3B model using our \gls{a:tool} dataset, we achieved a notable improvement in frame-level F1 performance, from 48.5\% to 87.5\%, achieving similar performance with GPT-4.1. 
Despite these results, the generalization capabilities of our model still trail behind state-of-the-art models such as GPT-4.1, particularly on Existential and Spatial queries. 

For future work, the following items are of immediate interest: (1) develop methods for synthetic generation of queries and subsequent perception streams to improve generalizability, (2) perform additional comparisons against a wider range of model configurations, and (3) incorporate other spatial information such as point clouds or depth maps.


\bibliographystyle{plain}
\bibliography{main}

\end{document}